\documentclass[letterpaper]{article}
\usepackage{aaai}
\usepackage{times}
\usepackage{helvet}
\usepackage{courier}
\usepackage{graphicx}
\usepackage{flushend}
\usepackage{url}

\begin{document}
\title{Intelligent Physiotherapy Through\\Procedural Content Generation}
%\author{Author 1 \& Author 2}
\author{Shabnam Sadeghi Esfahlani \& Tommy Thompson\\
Department of Computing \& Technology\\
Anglia Ruskin University\\
Cambridge, UK\\
shabnam.sadeghi-esfahlani@anglia.ac.uk, tommy@t2thompson.com
}

\maketitle
\begin{abstract}
\begin{quote}
This paper describes an avenue for artificial and computational intelligence techniques applied within games research to be deployed for purposes of physical therapy.  We provide an overview of prototypical research focussed on the application of motion sensor input devices and virtual reality equipment for rehabilitation of motor impairment: an issue typical of patients of traumatic brain injuries.  We highlight how advances in procedural content generation and player modelling can stimulate development in this area by improving quality of rehabilitation programmes and measuring patient performance.
\end{quote}
\end{abstract}

\section{Introduction}
In recent years computational intelligence (CI) and artificial intelligence (AI) have placed an increasing focus on applications within games. This research has led to a number of sub-fields to emerge, notably procedural content generation (PCG): the algorithmic process of constructing artefacts of some functional or cosmetic value. PCG is often adopted as a generic term that encompasses a variety of procedures whose applications range from the creation of visual aesthetics to the more ambitious realms of automated game development~\cite{TogeliusSurvey}.  In each instance, there are often regulations or rules imposed upon the generative systems in play: ensuring that content respects some pre-defined expectations.  These regulations may be imposed due to the functional needs of the environment the content is crafted for, or due to bias' inherent either in the human designers creative process or the computational and mathematical models adopted for content creation~\cite{DagstuhlPCG}.  These rules have a subsequent impact on the generative space of the final system and the types of content we can build for this problem.  In essence, this can ultimately inhibit the spaces within which generation will occur.

It is within this flexibility that we explore an interesting problem space for PCG research in physiotherapy: physical rehabilitation programmes devised to aid patients in improving or regaining motor functions. Physiotherapy is adopted in a wide variety of circumstances, such as personal injury or disability ranging from a variety of mental or physical complications. In order to maximise the gains of these rehabilitation programmes, patients must be provided with bespoke care and training that is reflective both of the patients current circumstances, but of the goals and expectations of the patient and their therapist. 

The last 10 years has seen a significant increase in the use of `serious games'\footnote{A term adopted for games with a primary interest in areas such as education and/or training rather than entertainment.  The term is often attributed given the stigma associated with games as a medium.} for the purposes of therapy and training, be it physical or mental~\cite{halton2008virtual}.  In any case, we typically require scenarios within these simulations that cater towards the patients needs.  This process is highly reliant on the expert-knowledge of physicians in order to achieve or provide momentum towards a specific goal in mind for the patient.  Despite this expertise, there is tremendous risk associated with virtual training scenarios for physiotherapy, given the need to ensure it is scaled to the abilities of the patient - so as not to cause injury - but also with respect to the aspirations and goals of their therapy - thus ensuring they actually improve.  On a less critical note, there is the issue of maintaining a patients interest in the process and aiming to further their goals while introducing diversity into the training programme.

With these considerations in mind, this paper highlights prototypical procedural content generation research for physiotherapy; with a specific focus on the rehabilitation of stroke patients.  We highlight the problem area as well as existing research in the field of serious games for rehabilitation and the work to-date adopting AI and CI methods to stimulate these processes.  This is followed by an overview of the prototypical system devised and our current research in this area.  We conclude this paper with a brief discussion of how PCG combined with player modelling research can prove highly influential in games of this nature.

\section{Background \& Related Work}
A stroke occurs when the blood supply to a particular part of the brain is suddenly cut off, resulting in cells in the affected area of brain become damaged~\cite{DepartmentofHealth}.  While potentially fatal, there are over 1.2 million stroke survivors currently residing in the United Kingdom (UK) with approximately 152,000 new cases identified each year~\cite{StrokeAssociation}.  Stroke survivors suffer from a number of symptoms, but some of the most long-term damaging include reduced mobility and a lack of co-ordination and balance.  As a result, many stroke sufferers are the recipient of physiotherapy treatment in an effort to return to pre-stroke levels of mobility.  This is a costly and time consuming process estimated to cost the UK economy £9 billion a year~\cite{halton2008virtual}.  Regardless of economic factors, the success of this treatment is variable due to the nature of strokes themselves: with some patients requiring longer periods of physiotherapy before regaining their former independence.  In addition, the continued motivation of the patient in working towards their goals~\cite{DepartmentofHealth} is equally important. 

In recent years, Virtual Reality (VR) has been identified as having potential therapeutic benefits~\cite{halton2008virtual}. Through use of VR technology, clinicians can provide an environment to practice and monitor repetitive motions and provide feedback to guide the rehabilitation process.  This has been found to improve rehabilitation for specific motor-learning tasks of patients suffering from traumatic injuries~\cite{ustinova2014virtual}.  This includes the movement of hands, upper and lower limbs~\cite{carlozzi2013using,saposnik2010effectiveness}, unilateral and bilateral movement~\cite{mirelman2009effects}.  Given that these platforms can range from high-cost bespoke systems to low-cost off-the-shelf video gaming technologies, it presents an opportunity - albeit with risk - for patients to perform additional exercises at home without the supervision of the therapist~\cite{koenig2014introduction}.  Lastly, it allows patients to focus on the training process given the immersive nature of the simulated environment \cite{dobkin2004strategies,holden2002}.  

In many respects the ideas of rehabilitation-driven games share similarities with that of dynamic difficulty adjustment.  However, as noted in the likes of~\cite{burke2009optimising}, the challenges faced are more nuanced given the physical challenges faced by patients.  A notable example can be found in the Intelligence Game Engine for Rehabilitation (IGER) detailed in~\cite{pirovano2012self,borghese2013computational} which is focussed on the recording and management of physiotherapy for stroke patients.  This system adopts an adaptive fuzzy-driven approach to catering for the patients rehabilitation process.  This is achieved while working with a variety of small mini-games that utilise peripherals such as the Microsoft Kinect and Nintendo Wii Fit board.  This system places emphasis on ensuring that the parameters of the current game are such that it should not result in activities that player might find painful.

\section{Project Overview}
\label{sec:prototype}
\begin{figure}
\includegraphics[width=0.5\textwidth]{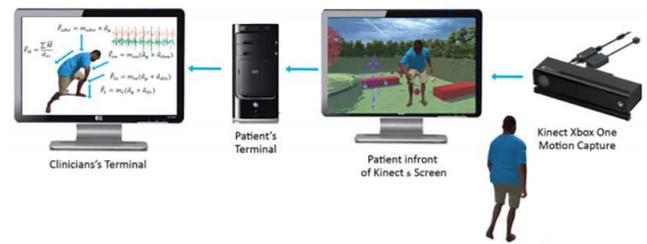}
\caption{\label{fig:system}A summary of the current prototype system: in which patients can interact with a game-environment courtesy of the Kinect interfaces.}
\end{figure}

The prototype gameplay system shown in Figure~\ref{fig:system} is designed to aid stroke patients to stimulate motor movement.  While our long-term efforts seek to adopt the use of virtual reality headsets\footnote{VR headsets are already implemented for this system, but is beyond the scope of current research interests}, our initial testing - and focus of this paper - is working solely with gesture recognition apparatus: namely the Microsoft `Kinect' camera and `Myo' wearable technologies~\footnote{\url{http://www.myo.com}}.  The Kinect camera allows for basic gesture expressions to be captured with Kinect, while the Myo provides more accurate gyroscopic data in addition to muscular activity.  This allows for movements to be monitored and co-ordination abnormalities to be recognised through subsequent analysis at a rate of approx. 30 frames per second.  

\begin{figure}
\includegraphics[width=0.5\textwidth]{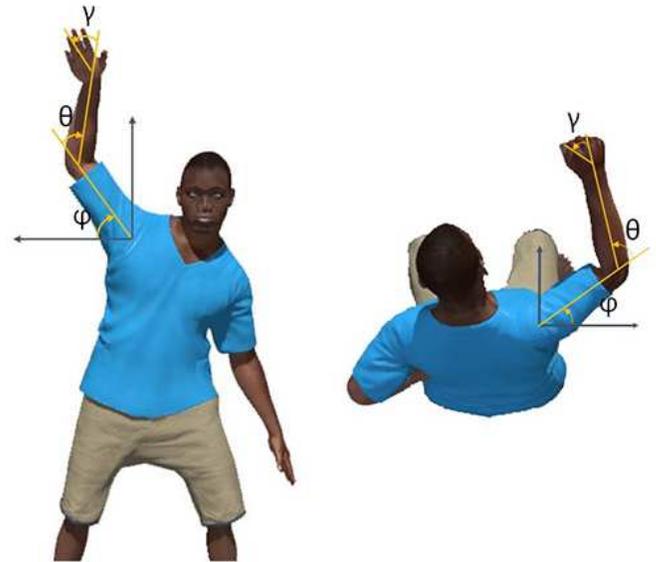}
\caption{\label{fig:kinect}The Kinect and Myo sensors allow for accurate representation of the horizontal and vertical motion of the player.}
\end{figure}

\begin{figure}
\includegraphics[width=0.48\textwidth]{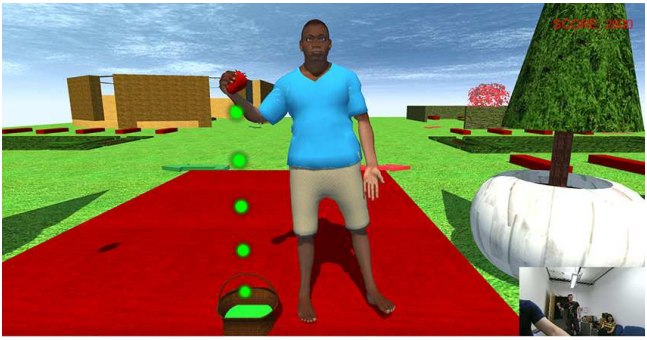}
\caption{\label{fig:Gameplay}A screenshot of the current working prototype, with an emphasis on interaction with objects in the virtual 3D space.}
\end{figure}

The adoption of Kinect and Myo allows us not only to model the patient within the scene but enables full-body interaction including upper and lower limbs (Figure~\ref{fig:kinect}). Through these sensors, the participant is able to manipulate the behaviour of a player (avatar) within the environment as shown in Figure~\ref{fig:kinect}. Subsequent therapy sessions are presented using a series of mini-games that require specific kinematic activity such as that shown in Figure~\ref{fig:Gameplay}.  At the time of publication, two simple games are implemented.  The first is a button-pressing game where players must hover their hand in the location to enable them.  The second is a fruit grabbing game (Figure~\ref{fig:Gameplay}) where players must grab objects and drop them in the basket at their feet.  These games only requires simple single-limb guided movements.  However, future plans aim to adopt complex, whole-body task-oriented actions such as achieving specific posture, balance, eye-tracking and facial expression tasks.

With these components in place, a control system has been established for the purposes of conducting physical rehabilitation exercises.  However, it is the usage of patient data both in terms of current kinematic and muscular performance and a developed understanding of the needs of the patient could yield a more comprehensive and reactive physiotherapy programme.  It is at this juncture we argue that computational intelligence methods found within games research that could prove useful in elevating this work.

Our ultimate goal is for both patients and physicians to interact with the system.  While patients continue to play these mini-games and can see their progress against increasingly challenging tasks, physicians can assess the progress on a more granular and data-driven level.  This will help identify ranges of motion that players have either not yet fully explored or have not performed well within, allowing for the physician to then tailor the system parameters to move towards more specific goals in-line with their expectations of the patient.

\section{Discussion}
The intention of this prototype is to enhance stroke therapy in an engaging and safe fashion.  Our aim is to allow participants with different movement capabilities to play the game and at the same time minimize any frustration.  Furthermore, we wish to address the possibility of providing patients with real-time feedback to enable correction and adaptation of erroneous behaviour.  This presents a rather interesting opportunity for procedural content generation.  

The inputs as detailed in Figure~\ref{fig:kinect} translates to a large range of motion.  However, this space is in-turn limited by the actual capabilities of the player, assuming they are suffering from limitations in mobility.  As such, games previously implemented such as the aforementioned button pressing or fruit grabbing are - at the time of writing - not fit for purpose.  No constraints are exhibited upon them beyond a range of possible physical space that a player can interact within.  They do not take into consideration the perceived and desired mobility of the player.  This leaves us with numerous challenges to face, which we will now briefly summarise in addition to some initial thoughts on tackling them.

\subsection{Perceived and desired mobility}
At minimum there are considerations to be made for generating within a restricted area of the player: understanding the range of comfortable motion within which the player can move should be used to automatically constrain the generative system with respect to the physical placement while minimising the impact upon expressivity of the generation itself~\cite{TogeliusSurvey}.  There is precedent for the adoption of parametrisation in order for players to dictate generated content, most notably the Mario AI level generation track~\cite{shaker20112010}: in which data acquired from a preliminary playthrough is adopted as a basis for subsequent level creations.  While these parameters are based on player performance in the case of~\textit{Super Mario}, we consider the range of movement exhibited by players in preliminary 'warm-up' games as means to manage future behaviour.  The subsequent generation can be tackled through use of vanilla genetic algorithms to isolate potential parameters for object placement, with constraints adopted as heuristics.  This could resulting in a variety of levels for games - or even mini-games themselves - that constrain themselves with respect to the perceived motion of the patient.  This would help not only maintain an element of novelty in the terms of player engagement, but maximise the range of motions being adopted by the patient.

\subsection{Physicians Expectations}
There is an interesting opportunity in the adoption of player-driven metrics in this problem space, given that the `levels' or tasks crafted for a particular play session should be geared towards not just the players current ability, but also the expectations of their physician.  Furthermore, there is the additional consideration of ensuring that the games crafted also challenge these actual and perceived limitations in such a fashion that it results in continued gains.  One possibility is to measure and curate levels/games based on how well they may suit specific tasks.  This is critical for it to prove practical as part of the physiotherapy process, given it would help maintain a critical focus for future therapy sessions.

\subsection{The Risk of Injury}
A crucial aspect of this research is minimising the risk of injury.  Unlike most procedural and adaptive play domains, we are dealing with circumstances that are safety-critical in nature and any interaction with the system could prove equally as detrimental to a patient as it could be beneficial.  As such, constraints imposed upon the generative system must ensure any risk presented towards players is minimised.  This is critical given not only the need to ensure no harm comes to the patient, but to retain their trust in a systems aimed at assisting them in their development. While a more robust means to achieve this is yet to be established, the formative 'practice-run' mentioned earlier would prove useful at establishing just how far a player will move in these sessions: given that patients will typically not deliberately cause themselves injury.

\subsection{Patients Perceptions of Ability}
Having previously discussed the issue of managing physicians expectations, there are also the challenges in addressing the physical and mental limitations of the patient.  Players may err on the side of caution for fear of injury, which will subsequently impede their development.  There are two issues to consider in this regard: the perception of whether the generated content could prove harmful to the user and whether the user believes the content is beyond their ability.  The modelling of such qualitative statements in reaction to the specified metrics has previously been tackled through the likes of fuzzy rules~\cite{pirovano2012self} as well as neuro-evolutionary preference learning~\cite{shaker2011feature}.  In each case, levels are qualified through qualitative metrics: an approach that can not only establish perceived difficulty, but a similar approach can be adopted for expectations of therapists.  
By training systems to establish what is feasible versus what is `achievable' in the minds of players, we can isolate problem areas where players may still gain tangible improvement.

\section{Conclusion}
This paper highlights early-stage research in the development of a game-driven framework for physical therapy and rehabilitation: highlighting the potential problem areas and challenges faced when developing generative systems that act within a safety-critical system.  This work is still in its formative stage, with the initial phase focussed on players providing initial data to aid in configuring subsequent level generation currently implemented, with future work focussed on establishing models of perceived difficulty for generative systems to tailor games for individual players.

\bibliography{Ref}
\bibliographystyle{aaai}

\end{document}